%% file: iclr2023_conference.tex
\documentclass{article} 
\usepackage{iclr2023_conference,times}

\input{math_commands.tex}

\usepackage{hyperref}
\usepackage{url}

\usepackage{graphicx}

\title{JPEG Compressed Images Can Bypass Protections Against AI Editing}

\author{Pedro~Sandoval-Segura, Jonas Geiping, Tom Goldstein\\
Department of Computer Science\\
University of Maryland, College Park\\
\texttt{\{psando, jgeiping, tomg\}@cs.umd.edu} \\
}

\newcommand{\eg}{\textit{e}.\textit{g}.}

\iclrfinalcopy 
\begin{document}

\maketitle

\begin{abstract}
Recently developed text-to-image diffusion models make it easy to edit or create high-quality images. Their ease of use has raised concerns about the potential for malicious editing or deepfake creation. Imperceptible perturbations have been proposed as a means of protecting images from malicious editing by preventing diffusion models from generating realistic images. However, we find that the aforementioned perturbations are not robust to JPEG compression, which poses a major weakness because of the common usage and availability of JPEG. We discuss the importance of robustness for additive imperceptible perturbations and encourage alternative approaches to protect images against editing.

\end{abstract}

\section{Introduction}

Given a user-defined text description of an image, diffusion models \citep{ho2020denoising, rombach2022high} excel at generating high-quality image samples. Diffusion models can also be used to edit specific parts of an existing image, guided by a text-prompt \citep{rombach2022high}. Previously, photo editing software and expertise was required to make serious scene adjustments or object additions, but now it can be done with a simple natural language phrase. The barrier to creating new images is relatively low, opening the door to malicious actors who could edit images or create deepfakes with a single click. To prevent the malicious editing of images, recent work has proposed using imperceptible perturbations to protect images from being modified \citep{salman2023raising, shan2023glaze}. As with other methods which modify images to prevent a deep learning model from accomplishing a task, releasing protected images online exposes the owner to a static defense vs. adaptive attack scenario \citep{radiya-dixit2021data}. In this setting, the protected images are required to withstand adaptive attacks from current methods and all future methods, a nearly impossible requirement. By successfully editing protected images, we demonstrate that JPEG compression remains a necessary baseline for protection methods that use imperceptible noise. The widespread use of JPEG transformations \citep{w3techs}, which may occur inadvertently (\eg, if JPEG is the default file format for capturing screenshots), suggests that optimized perturbations for image protection are exceedingly fragile. The robustness of perturbations is especially important in situations where image misuse by adversaries if of concern. All experiments are conducted using the open-source notebooks from the photoguard repository \citep{salman2023raising}, which employs the Stable Diffusion Model (SDM) v1.5 \citep{rombach2022high}.


\subsection{photoguard: Protecting Against Malicious AI-Powered Image Editing}
\label{subsection:photoguard-overview}

\citet{salman2023raising} propose methods for protecting images against manipulation by Stable Diffusion \citep{rombach2022high}. Their software, \textit{photoguard}, protects images against manipulation by adding an imperceptible adversarial perturbation designed to force a target diffusion model to generate unrealistic images. The authors propose two different protection techniques: an encoder attack and a diffusion attack. 

\textbf{Encoder attack}. The goal is to find an imperceptible perturbation $\delta_{\rm encoder}$ that, when added to the original image, results in a protected image that the SDM considers similar to a gray target image: 

\begin{equation}
    \delta_{\rm encoder} = \argmin_{\|\delta\|_{\infty} \leq \epsilon} \|E(x + \delta) - z_{\rm target} \|_{2}^{2}
    \label{eq:encoder-attack}
\end{equation}

\noindent
where $x$ is the image to be protected, $z_{\rm target}$ is some target latent representation, and $E$ is the image encoder. For a $512 \times 512 \times 3$ dimensional image, $\epsilon = 0.1$ and 40 steps of PGD with step-size $\frac{0.1}{40} \times 6$ is used to solve the optimization.

\textbf{Diffusion attack}. The goal is to find an imperceptible perturbation $\delta_{\rm diffusion}$ that, when added to the original image, results in a protected image that targets the diffusion process itself.
\begin{equation}
    \delta_{\rm diffusion} = \argmin_{\|\delta\|_{\infty} \leq \epsilon} \|f(x + \delta) - x_{\rm target} \|_{2}^{2}
    \label{eq:diffusion-attack}
\end{equation}
where $f$ is the SDM, $x$ is the image to be protected, $x_{\rm target}$ is the target image to be generated. Eq.~\ref{eq:diffusion-attack} is solved approximately using PGD, backpropagating through only $4$ diffusion steps due to GPU memory requirements \citep{salman2023raising}.

\input{figures/overview-figure-img2img}

\cite{salman2023raising} note that photoguard's effectiveness may diminish when protected images undergo transformations, and propose methods for making the imperceptible perturbation more robust \citep{athalye2018synthesizing}. Protecting images using an imperceptible perturbation optimized using either of the attacks of Section~\ref{subsection:photoguard-overview} is a interesting and novel contribution, but our research has found it is more brittle than expected to JPEG distortion, which poses a major weakness due to the common usage and availability of JPEG. Previous work has shown that a differentiable JPEG module can be added to the loss function to create JPEG-resistant adversarial examples \citep{shin2017jpeg}. One can imagine making a similar, simple modification to Eq.~\ref{eq:encoder-attack} and Eq.~\ref{eq:diffusion-attack} so that the perturbations are JPEG-resistant. But the question remains: how many times does one want to continue the back and forth between the data protector and adversary?



\subsection{Glaze: Protecting Against Style Mimicry}

\citet{shan2023glaze} propose a method to prevent fine-tuned diffusion models from learning a particular artist's style. In essence, imperceptible perturbations are optimized so that SDM finetuning on protected artist images, in the manner described by DreamBooth \citep{ruiz2022dreambooth}, is unsuccessful. In this way, new images generated by the finetuned model are unable to replicate the style captured in the protected subset of artist images. The proposed solution is unique and the authors perform experiments to test the robustness of their perturbations to Gaussian noise and JPEG image compression. However, only three compression levels between $20$ and $10$ are considered. Our experiments suggest that considering more subtle compression levels (\eg, between compression levels $95$ and $65$) are necessary. Excessive compression, even without the use of optimized imperceptible perturbations, could potentially damage the image to the extent that the style appears to not have been replicated.

Evaluating whether generated images replicate a particular style, or whether generated content satisfies some other criteria in general, presents a challenge due to its subjective nature. Our experiments suggest that imperceptible perturbations must possess greater robustness against a wider range of image transformations than those commonly considered, as final transformations could be made by a malicious actor.

\section{Bypassing Photoguard Protection}

At a high level, photoguard adds imperceptible noise to an image so that the image encoder in a diffusion model is unable to encode the image reasonably, resulting in downstream error by the denoising network. The idea being that if an adversary can only produce unrealistic images, they would choose not to edit the protected image. In this section, we demonstrate that if the adversary JPEG compresses the photoguard-protected image, they can effectively perform edits using the same SDM used to protect the image. We also discuss the possibility of optimizing against this attack and analyze how other simple image transforms affect photoguard protection.

\input{figures/jpeg-compression-img2img-figure}

\subsection{JPEG Compression as a Low-Cost Transform}
Given a source image (\eg, an image of a dog in the grass) and a text prompt (\eg, ``dog under heavy rain and muddy ground real''), SDM can edit the source image to match the prompt. In Figure~\ref{fig:img2img-overview}, we show that when the photoguard-protected image (from the Encoder attack) is used as the source, the edited image no longer maintains the original subject and the dog's front paw seems to blend with some kind of foam. However, if we JPEG compress the photoguard image with quality of 65, the edited image looks like the edit from the original, and is potentially even more faithful to the text prompt due to the more prominent rain.

To perform inpainting, SDM is conditioned on both a source image and a mask indicating the regions of the image that are to remain unedited. In the last row of Figure~\ref{fig:inpainting-overview}, we show that an adversary can take a JPEG compressed photoguard image (from the Diffusion attack) and generate a new image that contains more realistic background and clothing. 

\input{figures/overview-figure-inpainting}
\input{figures/jpeg-compression-inpainting-figure}

We explore the effect that JPEG compression of varying quality has on the resulting edited image in Figure~\ref{figure-jpeg-compression-img2img} for the Encoder attack and Figure~\ref{figure-jpeg-compression-inpainting} for the Diffusion attack, finding that there exists a compression level range in which image editing performs well. If the photoguard perturbations were mostly high-frequency, using a Gaussian blur should also be effective at restoring an adversary's editing abilities. However, in Appendix~\ref{apdx:gaussian-median-blurs} we find that Gaussian and median blurs are ineffective. Other simple image transforms like rotations and flips (Appendix~\ref{apdx:rotations-flip}) are also ineffective. This is likely because the image encoder is mostly invariant to these transforms. While these simpler transforms were unsuccessful, we believe they should still be part of extensive robustness evaluations of future methods utilizing imperceptible perturbations. JPEG's ability to undermine the protective perturbation highlights the transforms' uniqueness in eliminating the perturbation while maintaining important semantic content.

Prior work has shown that JPEG could be used as a reasonable pre-processing step to defend against adversarial examples \citep{das2017keeping}, but such pre-processing defenses are not as robust to adaptive attackers, which go last in the classical game of adversarial examples. As pointed out in \citet{salman2023raising}, this could imply that an approach like photoguard could be improved to be robust against these transformations. Yet, we want to highlight that the game between attack and defense is \textit{flipped} for data modifications, a modification like photoguard has to anticipate future adaptive attacks that scrub it, and previous results on robustness of adversarial attacks might not apply.


\section{Conclusion}

Ultimately, the attack presented in this note is surprisingly simple, to the point where the (re-) encoding with different JPEG settings could also be considered just a benign editing operation, or a standard operation when storing the modified image. This makes it harder to verify how high the bar is really raised by adversarial noise image protection methods. Our results suggest that an adversary can bypass protections by JPEG compressing images that have been imperceptibly modified for protection. Protecting images from misuse remains an extremely difficult problem. As long as semantic information of an image is recognizable by a human, as is the case with imperceptibly perturbed images, the human visual system is a constructive proof that there exists a function to recover the semantic information. Thus, for high-stakes settings where image misuse is of concern, imperceptible perturbations of any kind, regardless of the optimization details, are so far inadequate for protection against editing by an adversary.


\subsubsection*{Acknowledgments}
This work is funded in part by an Amazon Lab126 Diversity in Robotics and AI Fellowship. Additional funding is provided by DARPA GARD (HR00112020007).  

\bibliography{iclr2023_conference}
\bibliographystyle{iclr2023_conference}

\appendix
\section{Appendix}

\subsection{Gaussian and Median Blurs of the Protected Image}
\label{apdx:gaussian-median-blurs}

Gaussian blurs are low-pass filters which attenuate high-frequency signals. If the most important component of a photoguard perturbation were the high-frequency components, then a Gaussian blur should be successful at diminishing the protection. In Figure~\ref{figure-gaussian-blur}, we show Gaussian blurs of photoguard images do not enable SDM to generate reasonable images. Since Gaussian blurs also destroy a significant amount of semantic information, it is difficult to pinpoint whether the most important features of the photoguard perturbations are high-frequency. We also experiment with median blurs in Figure~\ref{figure-median-blur}.

\input{figures/gaussian-blur-figure}
\input{figures/median-blur-figure}

\subsection{Rotations and Horizontal Flip of the Protected Image}
\label{apdx:rotations-flip}

The adversary need not edit the photoguard-protected image as-is. Performing other simple transforms seems reasonable before editing the image. In Figures~\ref{figure-rotation} and \ref{figure-flip}, we experiment with rotating and flipping the protected image before using SDM. While the photoguard perturbation was not optimized so that rotations and flips induce the same embedding from the encoder, generated images did not retain the original subject of the source image, suggesting the photoguard perturbation was effective under these transforms.

\input{figures/rotation-figure}
\input{figures/flip-figure}

\end{document}

%% file: math_commands.tex
\usepackage{amsmath,amsfonts,bm}

\def\eqref#1{equation~\ref{#1}}

\def\1{\bm{1}}

\DeclareMathAlphabet{\mathsfit}{\encodingdefault}{\sfdefault}{m}{sl}
\SetMathAlphabet{\mathsfit}{bold}{\encodingdefault}{\sfdefault}{bx}{n}

\DeclareMathOperator*{\argmin}{arg\,min}

%% file: figures/overview-figure-img2img.tex
\begin{figure}[t]
\begin{center}
\includegraphics[width=0.7\textwidth]{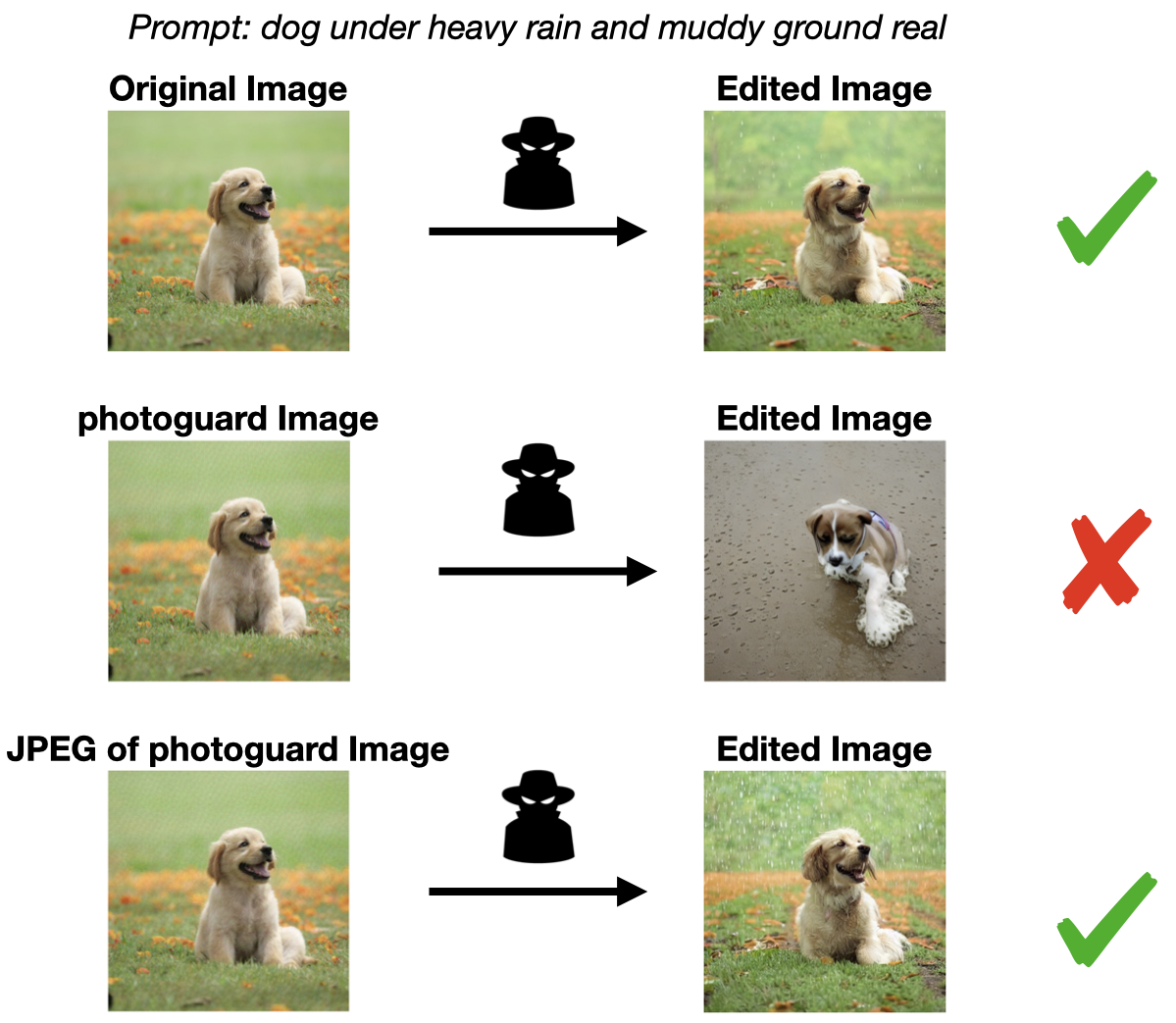}
\end{center}
\caption{\textbf{JPEG compression allows an adversary to modify a protected image found online.} First row: Given a text prompt, an adversary can make desired edits to an input image using a diffusion model. Second row: photoguard (Encoder attack) \citep{salman2023raising} protects the original image before an adversary can access it by adding an imperceptible perturbation. When the adversary edits the photoguard image, they are unable to maintain the original subject. Third row: By JPEG compressing the photoguard image, an adversary can edit the photoguard image while maintaining the original subject and adding key visual features of the text prompt.}
\label{fig:img2img-overview}
\end{figure}

%% file: figures/jpeg-compression-img2img-figure.tex
\begin{figure}[t]
\begin{center}
\includegraphics[width=0.75\textwidth]{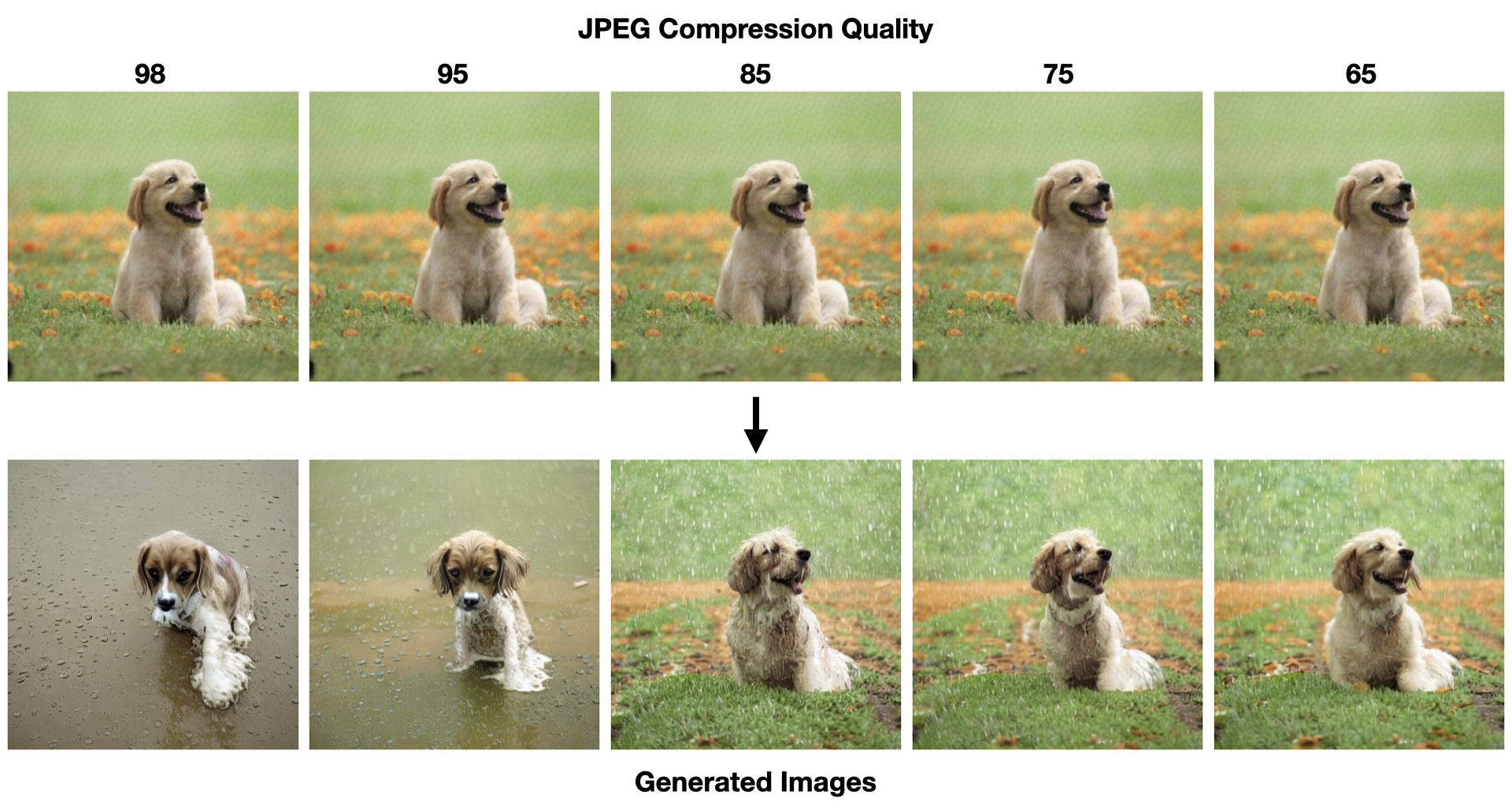}
\end{center}
\caption{\textbf{More JPEG compression more reliably undermines photoguard protection.} First row: We take a photoguard (Encoder attack) image  and JPEG compress it with varying quality. An image at 100\% JPEG quality is almost equivalent to the original photoguard image, while 65\% JPEG quality loses significant high-frequency information. Second row: Starting from the compressed image above, the adversary uses a diffusion model to make edits according to the same prompt setup as Figure~\ref{fig:img2img-overview}. With more compression, the generated content better maintains the original subject. Between compression quality of 95\% and 85\%, enough of the photoguard  noise is diminished, allowing stable image edits by an adversary.}
\label{figure-jpeg-compression-img2img}
\end{figure}

%% file: figures/overview-figure-inpainting.tex
\begin{figure}[t]
\begin{center}
\includegraphics[width=0.67\textwidth]{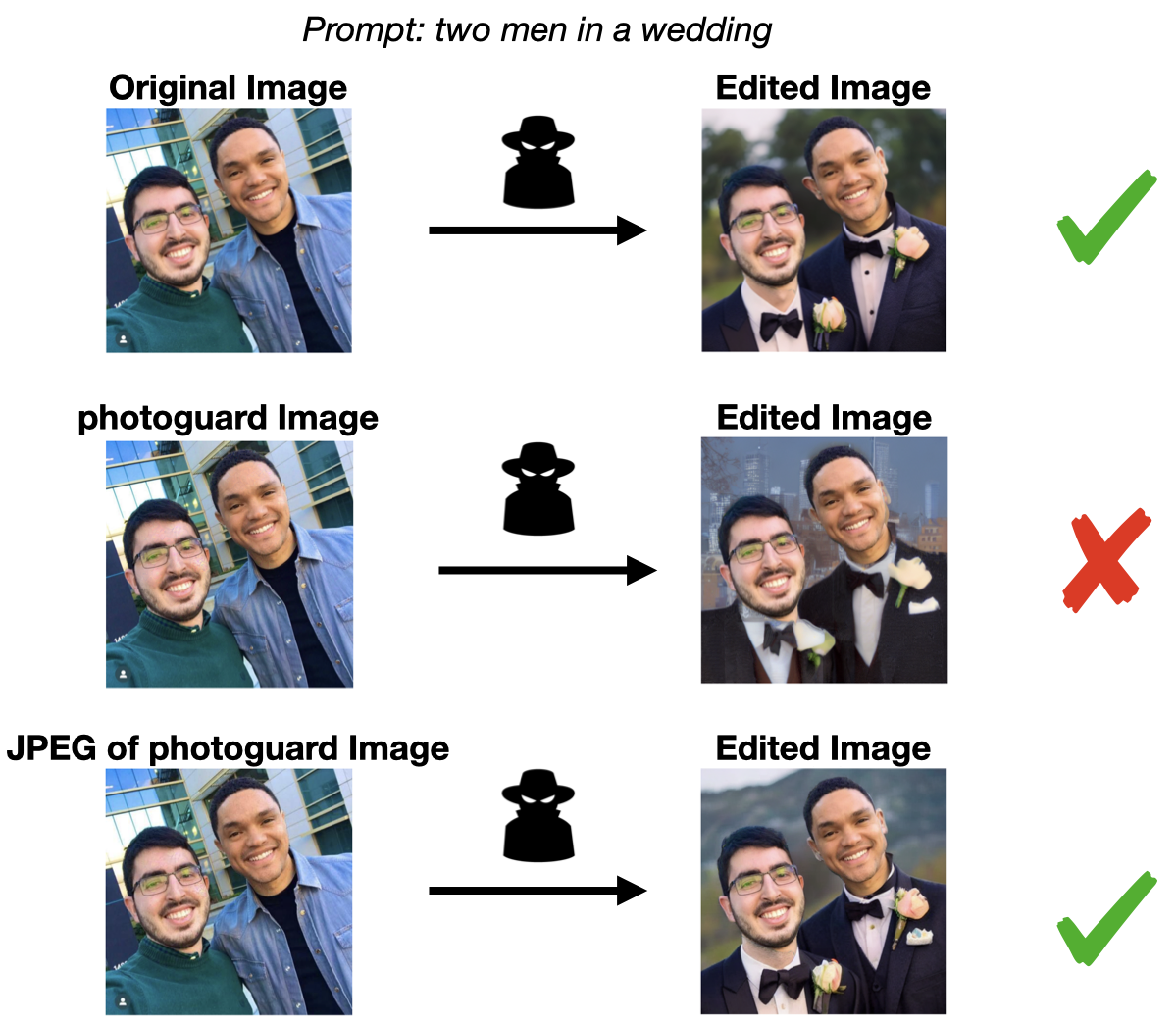}
\end{center}
\caption{\textbf{JPEG compression allows an adversary to edit the background of a protected image found online.} First row: Given a text prompt, an adversary can make desired edits to an input image using a diffusion model. Second row: photoguard (Diffusion attack) \citep{salman2023raising} protects the original image before an adversary can access it by adding an imperceptible perturbation. When the adversary edits the photoguard image, the background is unrealistic. Third row: By JPEG compressing the photoguard image, an adversary can edit the photoguard image while maintaining the original subjects and adding key visual features of the text prompt.}
\label{fig:inpainting-overview}
\end{figure}

%% file: figures/jpeg-compression-inpainting-figure.tex
\begin{figure}[t]
\begin{center}
\includegraphics[width=0.75\textwidth]{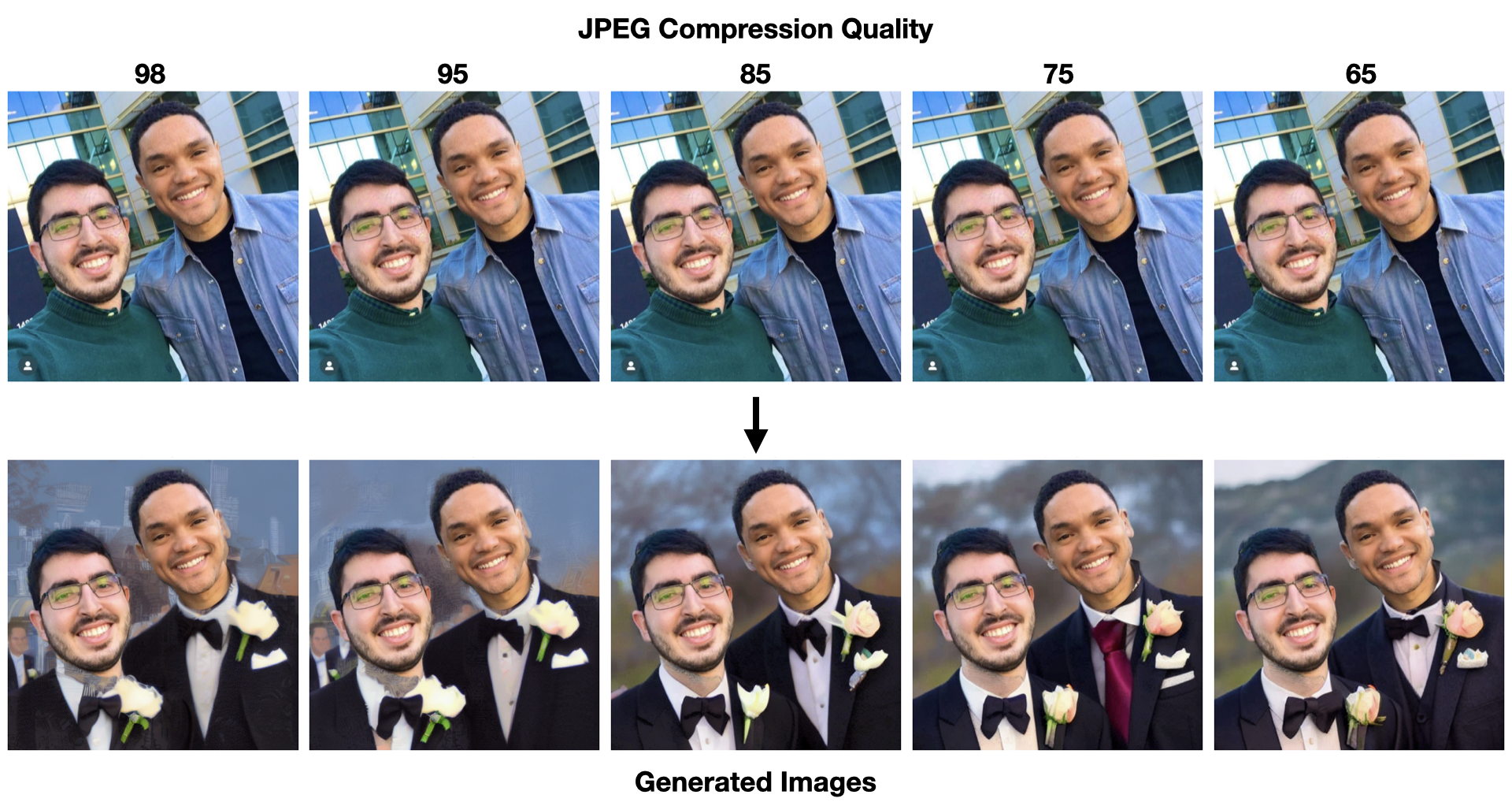}
\end{center}
\caption{\textbf{More JPEG compression more reliably undermines photoguard protection.} First row: We take a photoguard (Diffusion attack) image  and JPEG compress it with varying quality. Second row: Starting from the compressed image above, the adversary uses a diffusion model to make edits according to the same prompt setup as Figure~\ref{fig:inpainting-overview}. With more compression, the generated content background and clothing is more realistic. Between compression quality of 85\% and 75\%, enough of the photoguard  noise is diminished, allowing stable image edits by an adversary.}
\label{figure-jpeg-compression-inpainting}
\end{figure}

%% file: figures/gaussian-blur-figure.tex
\begin{figure}[ht]
\begin{center}
\includegraphics[width=0.55\textwidth]{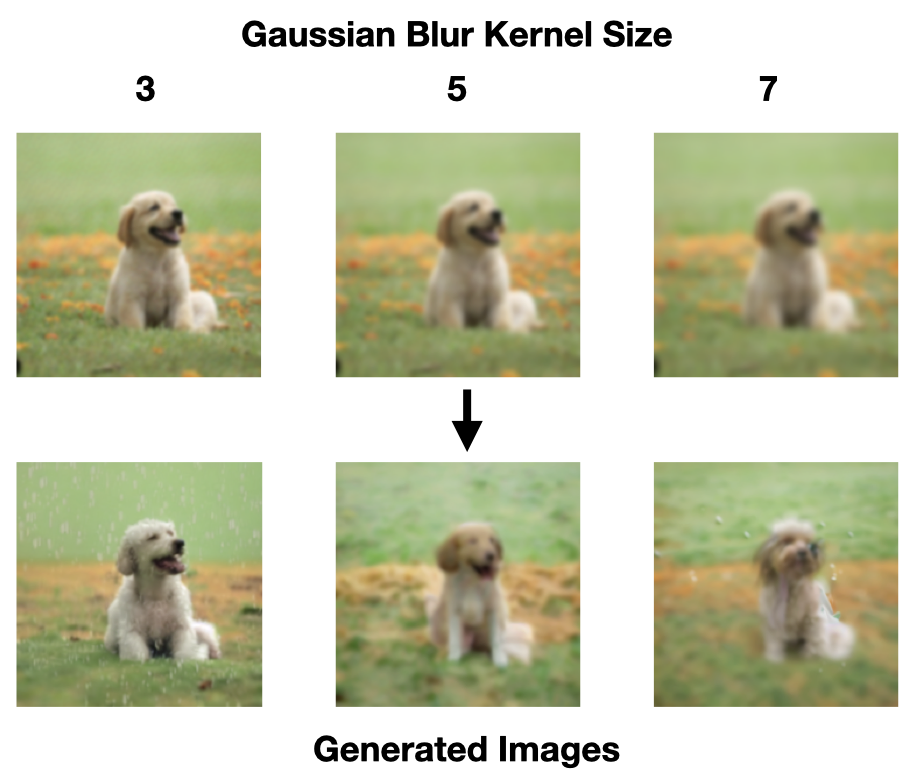}
\end{center}
\caption{\textbf{Gaussian blurs of photoguard images}. First row: We perform a Gaussian blur on a photoguard image with varying kernel sizes. Second row: Starting from the blurred image above, the adversary uses a diffusion model to make edits according to the same prompt setup as Figure~\ref{fig:img2img-overview}. With a larger kernel, the generated images lose significant visual features and do not maintain the original subject, unlike JPEG compression.}
\label{figure-gaussian-blur}
\end{figure}

%% file: figures/median-blur-figure.tex
\begin{figure}[h]
\begin{center}
\includegraphics[width=0.55\textwidth]{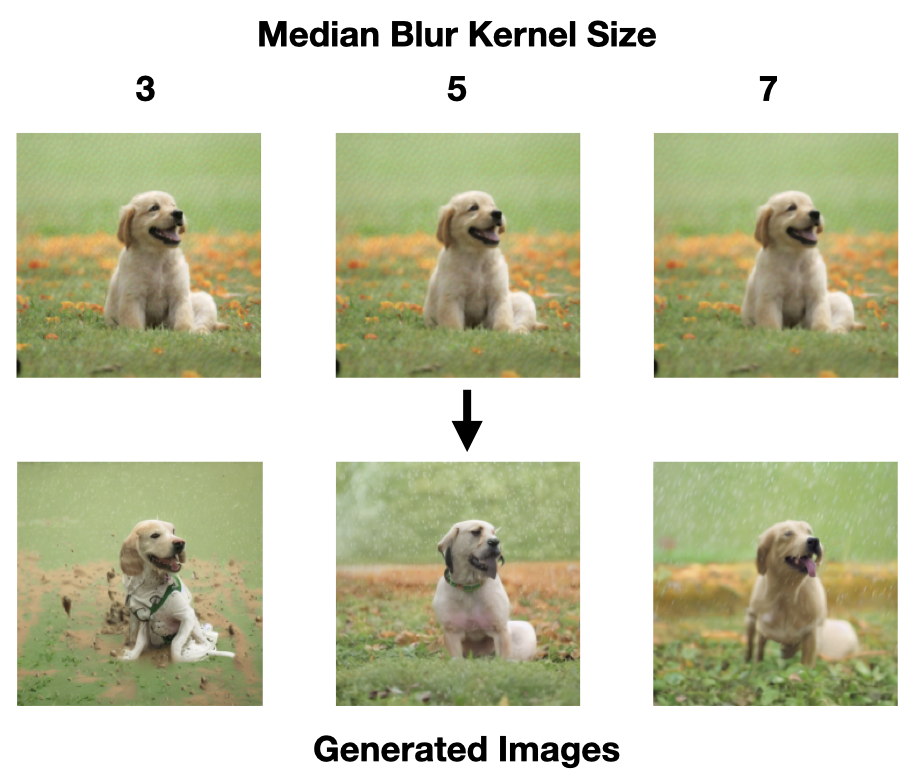}
\end{center}
\caption{\textbf{Median blurs of photoguard images}. First row: We perform a median blur on a photoguard image with varying kernel sizes. Second row: Starting from the blurred image above, the adversary uses a diffusion model to make edits according to the same prompt setup as Figure~\ref{fig:img2img-overview}. With a larger kernel, the generated images seemingly begin to retain the original subject.}
\label{figure-median-blur}
\end{figure}

%% file: figures/rotation-figure.tex
\begin{figure}[h]
\begin{center}
\includegraphics[width=\textwidth]{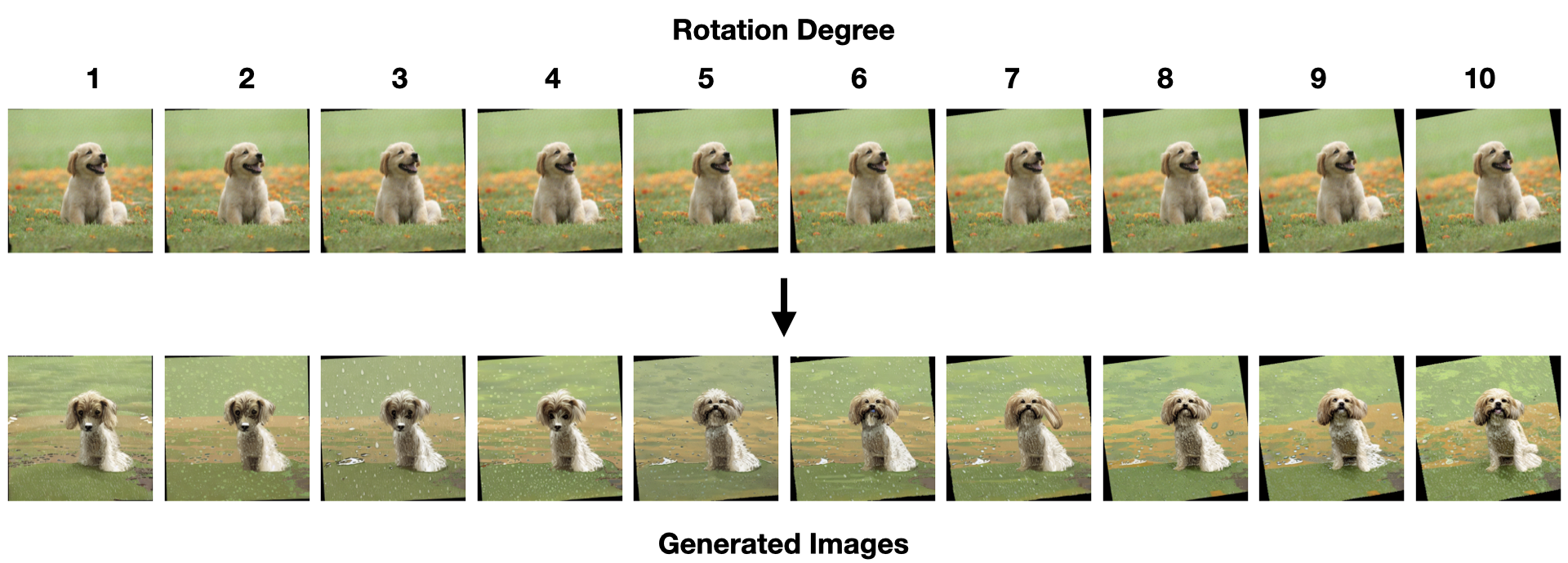}
\end{center}
\caption{\textbf{Rotations of photoguard images}. First row: We perform counter-clockwise rotations of varying angles on a photoguard image. Second row: Starting from the rotated image above, the adversary uses a diffusion model to make edits according to the same prompt setup as Figure~\ref{fig:img2img-overview}. The generated image does not retain the original subject in any of the rotations we tried.}
\label{figure-rotation}
\end{figure}

%% file: figures/flip-figure.tex
\begin{figure}[h]
\begin{center}
\includegraphics[width=0.58\textwidth]{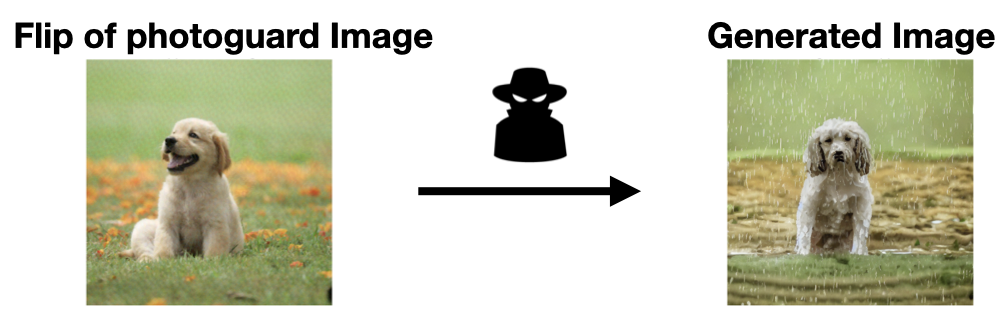}
\end{center}
\caption{\textbf{Flip of photoguard image}. Left: We perform a horizontal flip on a photoguard image. Right: Starting from the flipped image, the adversary uses a diffusion model to make edits according to the same prompt setup as Figure~\ref{fig:img2img-overview}. The generated image does not retain the original subject.}
\label{figure-flip}
\end{figure}